\documentclass[runningheads]{llncs}
\usepackage[T1]{fontenc}
\usepackage{graphicx,verbatim}

\usepackage{booktabs}  
\usepackage{amsmath}   
\usepackage{graphicx}  
\usepackage{xcolor}    
\usepackage{multirow}
\usepackage[colorlinks=true]{hyperref}  

\begin{document}


%
\title{All Centers Are at most a Few Tokens Apart: Knowledge Distillation with Domain Invariant Prompt Tuning}
\titlerunning{Knowledge Distillation with Domain Invariant Prompt Tuning}

%
\author{Amir Mohammad Ezzati \and
Alireza Malekhosseini \and
Armin Khosravi \and
Mohammad Hossein Rohban}
\authorrunning{A. Ezzati et al.}
%
\institute{
Department of Computer Engineering, Sharif University of Technology, Tehran, Iran\\
\email{iamirezzati@gmail.com}, \email{alrezmlk@gmail.com}, \email{arminkhosravie@gmail.com}, \email{rohban@sharif.edu}}



\maketitle              

\begin{abstract}
Domain generalization is critical in computational pathology (CPath) due to inherent domain shifts caused by variations in staining protocols, scanner devices, and imaging settings across clinical centers. Vision-language models (VLMs), such as PLIP—a pathology-tuned CLIP—trained on image-text pairs across diverse domains, serve as strong knowledge distillation sources. However, their zero-shot performance with predefined prompts remains limited due to sensitivity to prompt variations. Moreover, unlike natural images, histopathology centers lack semantic descriptors (e.g., 'sketch'), making it difficult to define domain-specific prompts for clinical centers. This requires a data-driven approach for learning domain-specific and  ultimately class-generic continuous prompts. We propose Domain Invariant Prompt Tuning (DIPT) for knowledge distillation process, a novel step that learns multiple input tokens for each domain. These tokens are trained separately for each domain and are averaged across domains, leading to domain-invariant prompts. Our student model then distills knowledge from PLIP’s text encoder by leveraging the prompts learned by DIPT. This leads to alignment of visual features with domain-invariant embeddings, enhancing generalization by training on multiple domains. Our method adds a significant improvement in average F1-score to existing state-of-the-art (SOTA) knowledge distillation approaches in domain generalization with histopathology datasets. This work helps the way of deploying robust CPath models in real-world clinical problems with heterogeneous data sources. The code is available at \href{https://github.com/amirezzati/dipt}{github.com/amirezzati/dipt}.


\end{abstract}

\section{Introduction}

Deep learning has demonstrated exceptional effectiveness in Computational Pat\-hology (CPath), enabling accurate histology image classification. However, domain shift reduces model generalization to unseen datasets, necessitating domain generalization (DG) solutions~\cite{jahanifar2023domain}.

Domain shift is a critical challenge in digital pathology, arising from differences in slide preparation, staining protocols, and scanner properties across medical centers. Even within the same institution, changes in the imaging pipeline over time can affect domain statistics. This issue results in deep learning models performing well on the training distribution but significantly worse on unseen domains~\cite{stacke2020measuring,van2021deep}.

Vision-language models (VLMs) like CLIP~\cite{radford2021learning} have recently shown great promise in enhancing out-of-distribution/domain generalization in the zero-shot setup~\cite{fang2022data}. They have demonstrated adaptability to various downstream tasks by learning joint embeddings of images and text through contrastive learning~\cite{radford2021learning}. Various follow up methods tried to enhance these models further through few-shot adaptation, showing remarkable robustness to domain drift~\cite{shakeri2024few}.

Despite all these advancements, CLIP is trained mostly on natural images, limiting its applicability in the domain of medical images. To address this challenge, Pathology language–image pretraining (PLIP), a CLIP-based model fine-tuned on OpenPath dataset, was introduced. It demonstrates strong zero-shot performance on unseen pathology domains corresponding to imaging centers~\cite{huang2023visual}.

 While pathology-pretrained models such as PLIP demonstrate remarkable zero-shot generalization across diverse domains, their absolute accuracy particularly for novel, specialized CPath tasks, fails to meet the necessary thresholds required for clinical deployment. One recent approach for addressing this issue is by improving domain generalization through Knowledge Distillation (KD)~\cite{addepalli2024leveraging}. Specifically, while naive KD struggles to enhance out-of-distribution (OOD) generalization, integrating distillation from other modalities, such as text, into the student model improves generalization.

Aligned with this idea, some recent methods~\cite{addepalli2024leveraging,huang2023sentence} in natural images propose distilling VLMs’ text encoders into the vision student model to enhance domain generalization. This is based on assuming text representations vary only slightly across domains, and are more generalizable than image representations~\cite{huang2023sentence,addepalli2024leveraging}. These methods usually append some descriptor of the domain into the text input, e.g. ``art'' for an artistic domain to account for slight deviations from the class-generic  representation. 

However, applying such methods in CPath is challenging due to the absence of semantic descriptors (e.g., ``art'' or ``sketch'') for histopathological data and the fact that existing class-generic prompts are not well-defined for histopathological images. In CPath, domains usually correspond to various imaging centers, which cannot be described expressively similar to natural images. To address this, we introduce Domain Invariant Prompt Tuning (DIPT), a novel step that generates domain-invariant and domain-specific prompts, enabling more effective knowledge distillation from PLIP’s text encoder. 


Specifically, our method consists of two main steps. In the first step, we concat learnable tokens alongside a fixed aggregated generic token to capture unknown domain-specific knowledge. These learnable tokens are trained using prefix tuning, allowing them to adapt to domain-specific variations while retaining generalizable features. After training, we compute embeddings of all domain-specific prompts for each class by passing them through the PLIP text encoder. By aggregating these domain-specific embeddings for a given class, we derive a domain-invariant class-generic embedding, which serves as a more general representation across domains.

In the second step, we freeze these domain-invariant class-generic embeddings for all classes and apply a KD pipeline to adapt the student model to various domains. This is achieved by distilling knowledge from both class-generic embeddings and the PLIP image encoder, ensuring the student model learns a more robust and generalizable representation. Experimental results on various benchmarks demonstrate the advantage of our model over zero-shot PLIP and other baselines, highlighting its effectiveness in improving domain generalization for computational pathology. Our contributions can be summarized as follows:
\begin{itemize}
    \item We are the first to apply knowledge distillation from vision-language models like PLIP to enhance domain generalization in histopathology datasets.
    \item We introduce a novel step, DIPT, before knowledge distillation to generate domain-specific and well-defined class-generic prompts.
    \item We conduct extensive evaluations across various source and target domain combinations, comparing different knowledge distillation (KD) approaches on the \textbf{CAMELYON17-WILDS}~\cite{koh2021wilds} and \textbf{Kather19}~\cite{kather2019predicting} datasets, achieving improvements in the F1-score metric of up to \textbf{7.7\%} compared to baseline methods.
\end{itemize}

\section{Backgrounds}

\subsubsection{Distillation from Vision Language Models.}

Knowledge distillation (KD)~\cite{hinton2015distilling} transfers knowledge from a high-capacity teacher model to a smaller student model, enabling the student to mimic the teacher’s outputs efficiently. Early KD methods relied on soft labels from teacher logits, which improve performance over direct supervision~\cite{hinton2015distilling,luo2016face,wang2021embracing}.
Later, H.~Chen et al.\ introduced feature embedding distillation~\cite{chen2020learning}, which leverages embeddings rather than logits.


For vision-language models (VLMs), KD benefits from an additional source: the text encoder. Traditional KD focused on vision models, transferring visual representations to a student model. However,
recent research shows that text encoders provide richer, domain-invariant semantic knowledge compared to image embeddings.

This shift has led to a new KD paradigm in VLMs, where knowledge transfer extends beyond visual representations. Notably, RISE~\cite{huang2023sentence} and VL2V-ADiP~\cite{addepalli2024leveraging} leverage text encoders to enhance vision student models with more generalizable information.

Both methods employ a cosine similarity loss (see Eq.~\ref{eq:image_sim_loss}) to align student image embeddings with those of the teacher model, similar to traditional distillation methods. This encourages the student’s feature space to mimic the generalizable feature space of CLIP’s image encoder, which has been trained on large-scale datasets using contrastive learning.

\begin{equation}
\mathcal{L}_{\text{I}} = \sum_{(\mathbf{x}, \mathbf{y})}  \cos(f(\mathbf{x}), h_I(\mathbf{x})),
\label{eq:image_sim_loss}
\end{equation}


where \( f(\mathbf{\cdot}) \) and \( h_I(\mathbf{\cdot}) \) are the student and teacher image encoders, respectively, and \(\cos(., .)\) refers to the cosine similarity function.


RISE introduces a novel loss function that distills knowledge from CLIP’s text encoder. This loss (see Eq.~\ref{eq:abs_distance_loss}) measures the absolute distance between the student’s image embedding and a generic text representation from the teacher's text encoder, which is obtained by aggregating predefined template prompts across all domains. This alignment enhances knowledge transfer from CLIP, improving generalization.


\begin{equation}
\mathcal{L}_{\text{A}} = \sum_{(\mathbf{x}, \mathbf{y})} \sum_{\mathbf{i \in C}} \mathbf{1}_{[y=i]} \cdot \cos(f(\mathbf{x}), \mathbf{E}_{i}).
\label{eq:abs_distance_loss}
\end{equation}


Here, \( \mathbf{E}_{i} \) is a domain-invariant class-generic embedding for class \( i \), \( \mathbf{1} \) is an indicator function, and \( C \) denotes the set of all possible classes.



However, in CPath, there is no well-defined, class-generic prompt that can be directly used as \( \mathbf{E}_{i} \) in the learning process. Moreover, there is no domain-specific descriptor for each domain, as the domains are different medical imaging centers, making it difficult to construct a domain-specific prompt that could be aggregated into a class-generic prompt. 

\subsubsection{Prompt Tuning.}
Prompt tuning adapts pretrained VLMs like CLIP by introducing task-specific textual tokens. While hand-crafted prompts (e.g., “a photo of a [CLASS]”) enable zero-shot classification, they often lack the flexibility to adapt to task-specific details. Context Optimization (CoOp)~\cite{zhou2022learning} replaces hand-crafted prompts with learnable soft prompts, improving adaptability. To further enhance generalization, some approaches ~\cite{zhou2022conditional,zang2022unified} introduce image-conditioned prompts, where image features are integrated with trainable textual prompts. Additionally, Knowledge-Guided Context Optimization (KgCoOp)~\cite{yao2023visual} refines these techniques by addressing the problem of overfitting to seen classes. While CoOp boosts accuracy on seen classes, it often struggles with unseen classes, partly due to the gap between learned prompts and CLIP’s generic prompt embeddings. KgCoOp mitigates this issue by incorporating a knowledge-guided loss function, which ensures learned prompts retain general linguistic knowledge while being optimized for specific tasks.

\section{Proposed Method}
While ~\cite{yao2023visual} focuses on generalization to unseen classes, our goal is to improve generalization across unseen domains in CPath. To achieve this, we introduce DIPT (Domain-Invariant Prompt Tuning), a novel preparatory step before knowledge distillation. DIPT extends the pipeline of knowledge distillation by first generating {\it domain-specific} prompts through adaptation on the PLIP model.
These domain-specific prompts are then aggregated to form domain-invariant class-generic embeddings. Applying DIPT before PLIP knowledge distillation demonstrates stronger generalization and robustness compared to pure KD approaches.



\begin{figure}[t]
\centering
\includegraphics[width=\textwidth]{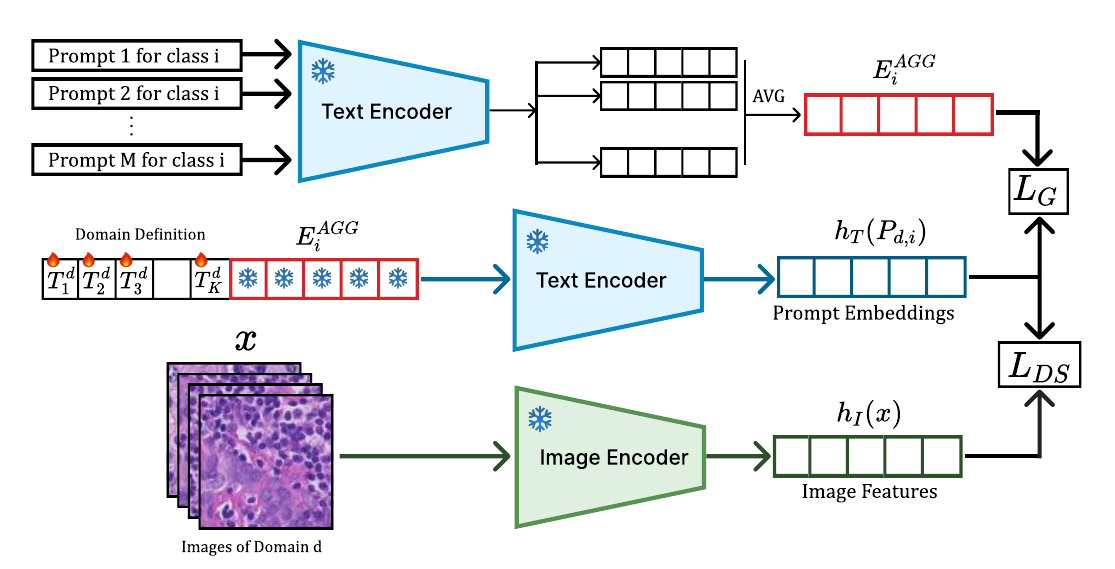}
\caption{\textbf{Domain-Specific Prompt Learning in DIPT.} Class-generic tokens are aggregated embeddings from multiple generic prompts per class (encoded via PLIP's text encoder). These tokens are concatenated with K learnable continous tokens used for adopting domain characteristics. Training these tokens adopts learnable prompts on PLIP for better classification.}

\label{fig:domain-spec}
\end{figure}

\begin{figure}[t]   
    \centering
    \includegraphics[width=\textwidth]{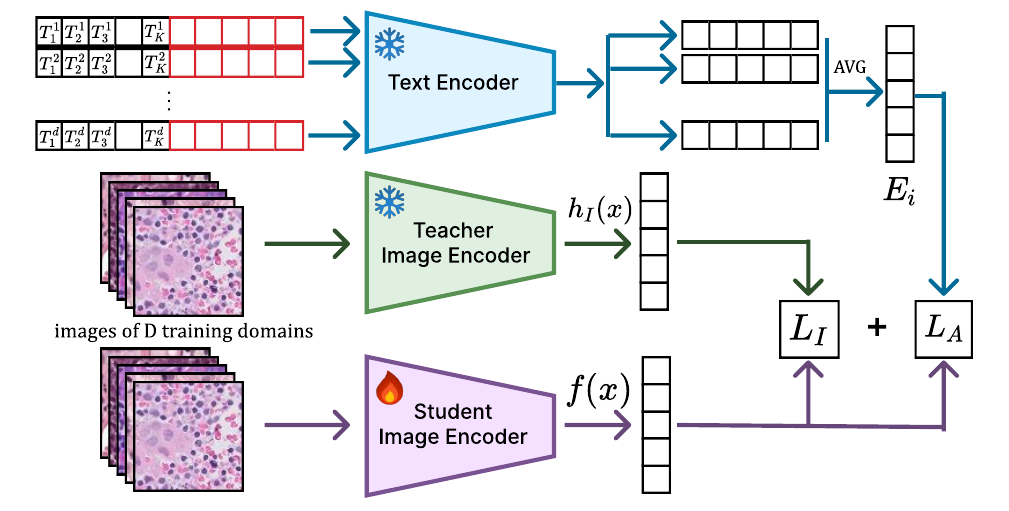}
    \caption{\textbf{Knowledge Distillation via Class-Generic Learned Prompts.} Domain-specific prompts from $D$ domains are aggregated after passing through PLIP's text encoder $h_T(\cdot)$ to form domain-invariant class-generic embeddings $\mathbf{E}_i$ (Eq.~\ref{eq:domain-inv}). These embeddings are then used for dual knowledge distillation.}
    \label{fig:Kd-pipeline}
\end{figure}

\subsection{Domain-Specific Prompt Learning}
The first step of DIPT i.e domain-specific prompt learning optimizes learnable tokens to capture domain-relevant features while maintaining generalizable knowledge through aggregated template embeddings. There are two types of tokens: domain-specific tokens and class-generic tokens, with the final prompt formed by their concatenation (as shown in Fig.~\ref{fig:domain-spec}).

Domain-specific tokens are \( k \) learnable vectors that adapt to domain-specific semantics. These tokens act as histopathological domain descriptors, which do not exist beforehand but are learned to represent domain characteristics such as staining and imaging device variations through feedback from the image encoder ($\mathcal{L}_{DS}$ loss in Fig. \ref{fig:domain-spec}). These tokens are initialized using a Gaussian distribution and updated during training.  


Class-generic tokens are frozen vectors that serve as a stable reference for each class and come from aggregated template embeddings. To construct these template embeddings, we generate \textbf{\( M \)} prompt templates for each class. For example, for the class ``normal lymph node,'' prompts like \textit{``a patch of normal lymph node''} and \textit{``benign lymphoid cells''} are used.
These prompts are encoded using PLIP’s text encoder, and their embeddings are averaged per class to form the final aggregated template embeddings.

Formally, for the \textbf{\( i \)}-th class, the aggregated template embedding \(\mathbf{E}_i^{Agg}\) is computed as:  
\begin{equation}
\mathbf{E}_i^{Agg} = \frac{1}{M} \sum_{m=1}^{M} h_T(\mathbf{P}_{i, m}^{Template}),
\end{equation}
where \(h_T(\cdot)\) is the PLIP text encoder, \(\mathbf{P}_{i, m}^{Template}\) is the \(m\)-th prompt template for class \(i\).
During training, domain-specific learnable tokens \( T_1^{d}, T_2^{d}, ..., T_k^{d} \) are optimized using a combined loss function that balances domain-specific discrimination and alignment with class-generic knowledge. The total loss (see Eq.~\ref{eq:total-loss}) integrates a standard cross-entropy term as domain-spesific loss: 



\begin{equation}
\mathcal{L}_{DS} = -\sum_{(\mathbf{x}, \mathbf{y})} \log \frac{\exp({z}_y / \tau) }{\sum_{j=1}^{N_c} \exp({z}_j / \tau)},
\label{eq:domain-spec-loss}
\end{equation}
where, 

\begin{equation}
{z}_i = \cos(h_I(\mathbf{x}), \mathbf{E}_{d,i}),
\label{eq:z-equation}
\end{equation}
and a generalization loss term:
\begin{equation}
\mathcal{L}_{G} = \frac{1}{N_c} \sum_{i=1}^{N_c} \cos(\mathbf{E}_{d,i}, \mathbf{E}_i^{Agg}),
\label{eq:gen-loss}
\end{equation}
where \( \mathbf{E}_{d,i} \) denotes the domain-specific prompt embedding for domain \( d \) and class \( i \). \( N_c \) is the total number of classes. Additionally, \( \tau \) is a temperature parameter,  and \( {z}_i \) represents the logit of class \( i \). These two losses are added to form the final loss:
\begin{equation}
\mathcal{L} = \mathcal{L}_{DS} + \mathcal{L}_{G}.
\label{eq:total-loss}
\end{equation}
\subsection{Knowledge Distillation via Class-Generic Learned Prompts} 
To improve cross-domain adaptability, we aggregate \textbf{\( D \)} domain-specific prompts learned in the previous step into a domain-invariant class-generic embedding (see Eq.~\ref{eq:domain-inv}). This enhances generalization by capturing stronger cross-domain semantics while reducing domain-specific biases, leading to better performance on both seen and unseen domains:
\begin{equation}
\mathbf{E}_i = \frac{1}{D} \sum_{d=1}^{D} h_T(\mathbf{P}_{d, i}),
\label{eq:domain-inv}
\end{equation}
where \(h_T(\cdot)\) is the PLIP's text encoder, \(\mathbf{P}_{d, i}\) is the domain-specific prompt for class \(i\) in the domain \(d\).

Finally, this class-generic embedding $\mathbf{E}_i$ is incorporated into the distillation pipeline (as shown in Fig.~\ref{fig:Kd-pipeline}). This process helps the model develop a broader understanding that extends beyond individual domains. By leveraging well-defined class-generic embeddings, the model enhances its ability to generalize across diverse medical imaging centers, improving generalization to unseen domains.

\begin{table}[]
\centering
\renewcommand{\arraystretch}{1.2}     
\setlength{\tabcolsep}{4pt}           
\caption{Test result on Camelyon17-WILDS. The results are reported with mean error bars, where the accuracy has a mean error of $\pm 0.18\%$ and the F1 score has a mean error of $\pm 0.21\%$. The subscript * indicates training with a ViT-based student, while others use ResNet-50. All reported values are in percent. (best results in \textbf{bold})}
\label{tab:comparison_domain}
\fontsize{8pt}{8pt}\selectfont  
\begin{tabular}{lcccccccc}
\toprule
\multirow{2}{*}{Method} & \multicolumn{2}{c}{Center 1} & \multicolumn{2}{c}{Center 3} & \multicolumn{2}{c}{Center 4} & \multicolumn{2}{c}{Center 5} \\
\cmidrule(lr){2-3}\cmidrule(lr){4-5}\cmidrule(lr){6-7}\cmidrule(lr){8-9}
 & ACC & F1 & ACC & F1 & ACC & F1 & ACC & F1 \\
\midrule

Agg. prompt Zero-Shot       & 81.89 & 81.79 & 89.38 & 89.64 & 77.63 & 77.27 & 75.58 & 79.98 \\
KD~\cite{hinton2015distilling}        & 93.94 & 93.61 & 88.57 & 87.39 & 91.42 & 90.75 & 83.32 & 80.20 \\
\midrule

RISE~\cite{huang2023sentence}      & 90.16 & 89.16 & \textbf{93.07} & \textbf{92.85} & 90.51 & 89.58 & 81.67 & 77.75 \\
RISE \textbf{+ DIPT}          & \textbf{95.98} & \textbf{95.87} & 91.23 & 90.51 & \textbf{94.13} & \textbf{93.84} & \textbf{83.37} & \textbf{80.14} \\
\midrule

VL2V~\cite{addepalli2024leveraging}       & 93.24 & 92.76 & 90.43 & 89.47 & \textbf{90.99} & \textbf{90.11} & 87.42 & 85.78 \\
VL2V \textbf{+ DIPT}          & \textbf{93.85} & \textbf{93.46} & \textbf{93.87} & \textbf{93.55} & 90.90 & 89.99 & \textbf{93.85} & \textbf{93.46} \\
\midrule

VL2V\textsuperscript{*}~\cite{addepalli2024leveraging}    & 96.32 & 96.31 & 93.81 & 93.49 & \textbf{95.24} & \textbf{95.06} & 88.66 & 87.23 \\
VL2V\textsuperscript{*} \textbf{+ DIPT}     & \textbf{96.85} & \textbf{96.86} & \textbf{96.45} & \textbf{96.40} & 94.47 & 94.18 & \textbf{93.66} & \textbf{93.28} \\
\bottomrule
\end{tabular}
\end{table}

\section{Experiments and Results}

\subsubsection{Datasets.}
The Camelyon17-WILDS dataset~\cite{koh2021wilds} consists of tissue patches from five different hospitals, focusing on breast cancer metastases in lymph nodes. Additionally, the Kather19 dataset~\cite{kather2019predicting} consists of colorectal tissue patches collected from three different centers, covering nine tissue types for histopathology analysis. They evaluate model generalization across unseen domains.

\subsubsection{Prompt Learning and Knowledge Distillation.}
In both steps, we reserve one of the centers (e.g. center 2 for Camelyon17) for validation. We experiment with \( k \in \{2,3,4\} \) and various learning rates (e.g. \( 5\times10^{-6} \), \( 5\times10^{-5} \)). For knowledge distillation, we evaluate VL2V (with ResNet-50 and ViT-B/16) and RISE (with ResNet-50) on Camelyon17 using a cross-domain setup that rotates the remaining four domains between training and test sets. 

\begin{table}[htbp]
\centering
\renewcommand{\arraystretch}{1.2}     
\setlength{\tabcolsep}{4pt}            

\begin{minipage}[t]{0.55\linewidth}
\centering
\fontsize{8pt}{8pt}\selectfont  
\caption{Mean and worst-case performance from Table~\ref{tab:comparison_domain} on Camelyon17 (best results in \textbf{bold}).}

\label{tab:comparison_mean_worst}
\begin{tabular}{lcccc}
\toprule
\multirow{2}{*}{Method} & \multicolumn{2}{c}{Mean} & \multicolumn{2}{c}{Worst} \\
\cmidrule(lr){2-3}\cmidrule(lr){4-5}
 & ACC & F1 & ACC & F1 \\
\midrule
Zero-Shot & 81.12 & 82.17 & 75.58 & 77.27 \\
KD~\cite{hinton2015distilling}         & 89.31 & 87.98 & 83.32 & 80.20 \\
\midrule
RISE~\cite{huang2023sentence}      & 88.85 & 87.33 & 81.67 & 77.75 \\
RISE \textbf{+ DIPT}     & \textbf{91.17} & \textbf{90.09} & \textbf{83.37} & \textbf{80.14} \\
\midrule
VL2V~\cite{addepalli2024leveraging}      & 90.52 & 89.53 & 87.42 & 85.78 \\
VL2V \textbf{+ DIPT}    & \textbf{93.11} & \textbf{92.61} & \textbf{90.90} & \textbf{89.99} \\
\midrule
VL2V\textsuperscript{*}~\cite{addepalli2024leveraging}      & 93.50 & 93.02 & 88.66 & 87.23 \\
VL2V\textsuperscript{*} \textbf{+ DIPT}     & \textbf{95.35} & \textbf{95.18} & \textbf{93.66} & \textbf{93.28} \\
\bottomrule
\end{tabular}
\end{minipage}\hfill
\begin{minipage}[t]{0.41\linewidth}
\centering
\fontsize{8pt}{8pt}\selectfont  
\caption{Test result comparison on Kather19 (best results in \textbf{bold}).}
\label{tab:comparison_domain0}
\begin{tabular}{lcc}
\toprule
Method & ACC & F1 \\
\midrule
Zero-Shot       & 63.66 & 60.86 \\
KD~\cite{hinton2015distilling}        & 92.87 & 92.98 \\
\midrule
RISE~\cite{huang2023sentence}      & 92.98 & 92.45 \\
RISE \textbf{+ DIPT}          & \textbf{93.73} & \textbf{93.75} \\
\midrule
VL2V~\cite{addepalli2024leveraging}    & 92.08 & 91.90 \\
VL2V\textbf{+ DIPT}     & \textbf{93.46} & \textbf{93.42} \\
\bottomrule
\end{tabular}
\end{minipage}
\end{table}

\subsubsection{Results.}
Table~\ref{tab:comparison_domain} summarizes the performance of baselines trained with DIPT-learned prompts. The original RISE and VL2V baselines are reported using an aggregated template embeddings and a single generic prompt, respectively, following their original implementations. On the Camelyon17-WILDS dataset, our method improves accuracy by up to \textbf{6.43\%}, while according to Table~\ref{tab:comparison_mean_worst}, the worst-case F1 score experiences an improvement of up to \textbf{6.05\%} (VL2V\textsuperscript{*}). Also the mean accuracy improves by up to \textbf{2.59\%}, and the mean F1 score increases by up to \textbf{3.08\%}, with consistent gains observed across different configurations. Similarly, on the Kather19 dataset, Table~\ref{tab:comparison_domain0} shows that combining our DIPT approach with KD methods further improves the test set F1 score by \textbf{1.52\%}.


\section{Conclusions}  
Our study shows that incorporating DIPT step before KD pipelines enhances domain generalization. DIPT entails learning and aggregating domain-specific  prompts in the embedding space. This aids in generation of a class-generic representation. Our approach improves model generalization and handles domain shifts in the histopathology, where domain definitions do not exist.

\bibliographystyle{splncs04}
\bibliography{references}

@article{jahanifar2023domain,
  title={Domain generalization in computational pathology: Survey and guidelines},
  author={Jahanifar, Mostafa and Raza, Manahil and Xu, Kesi and Vuong, Trinh and Jewsbury, Rob and Shephard, Adam and Zamanitajeddin, Neda and Kwak, Jin Tae and Raza, Shan E Ahmed and Minhas, Fayyaz and others},
  journal={arXiv preprint arXiv:2310.19656},
  year={2023}
}

@article{stacke2020measuring,
  title={Measuring domain shift for deep learning in histopathology},
  author={Stacke, Karin and Eilertsen, Gabriel and Unger, Jonas and Lundstr{\"o}m, Claes},
  journal={IEEE journal of biomedical and health informatics},
  volume={25},
  number={2},
  pages={325--336},
  year={2020},
  publisher={IEEE}
}

@article{van2021deep,
  title={Deep learning in histopathology: the path to the clinic},
  author={Van der Laak, Jeroen and Litjens, Geert and Ciompi, Francesco},
  journal={Nature medicine},
  volume={27},
  number={5},
  pages={775--784},
  year={2021},
  publisher={Nature Publishing Group US New York}
}

@inproceedings{shakeri2024few,
  title={Few-shot adaptation of medical vision-language models},
  author={Shakeri, Fereshteh and Huang, Yunshi and Silva-Rodr{\'\i}guez, Julio and Bahig, Houda and Tang, An and Dolz, Jose and Ben Ayed, Ismail},
  booktitle={International Conference on Medical Image Computing and Computer-Assisted Intervention},
  pages={553--563},
  year={2024},
  organization={Springer}
}

@article{huang2023visual,
  title={A visual--language foundation model for pathology image analysis using medical twitter},
  author={Huang, Zhi and Bianchi, Federico and Yuksekgonul, Mert and Montine, Thomas J and Zou, James},
  journal={Nature medicine},
  volume={29},
  number={9},
  pages={2307--2316},
  year={2023},
  publisher={Nature Publishing Group US New York}
}

@article{hinton2015distilling,
  title={Distilling the knowledge in a neural network},
  author={Hinton, Geoffrey and Vinyals, Oriol and Dean, Jeff},
  journal={arXiv preprint arXiv:1503.02531},
  year={2015}
}

@inproceedings{wang2021embracing,
  title={Embracing the dark knowledge: Domain generalization using regularized knowledge distillation},
  author={Wang, Yufei and Li, Haoliang and Chau, Lap-pui and Kot, Alex C},
  booktitle={Proceedings of the 29th ACM international conference on multimedia},
  pages={2595--2604},
  year={2021}
}

@inproceedings{luo2016face,
  title={Face model compression by distilling knowledge from neurons},
  author={Luo, Ping and Zhu, Zhenyao and Liu, Ziwei and Wang, Xiaogang and Tang, Xiaoou},
  booktitle={Proceedings of the AAAI conference on artificial intelligence},
  volume={30},
  number={1},
  year={2016}
}

@article{chen2020learning,
  title={Learning student networks via feature embedding},
  author={Chen, Hanting and Wang, Yunhe and Xu, Chang and Xu, Chao and Tao, Dacheng},
  journal={IEEE Transactions on Neural Networks and Learning Systems},
  volume={32},
  number={1},
  pages={25--35},
  year={2020},
  publisher={IEEE}
}

@inproceedings{huang2023sentence,
  title={A sentence speaks a thousand images: Domain generalization through distilling clip with language guidance},
  author={Huang, Zeyi and Zhou, Andy and Ling, Zijian and Cai, Mu and Wang, Haohan and Lee, Yong Jae},
  booktitle={Proceedings of the IEEE/CVF International Conference on Computer Vision},
  pages={11685--11695},
  year={2023}
}

@inproceedings{addepalli2024leveraging,
  title={Leveraging vision-language models for improving domain generalization in image classification},
  author={Addepalli, Sravanti and Asokan, Ashish Ramayee and Sharma, Lakshay and Babu, R Venkatesh},
  booktitle={Proceedings of the IEEE/CVF Conference on Computer Vision and Pattern Recognition},
  pages={23922--23932},
  year={2024}
}

@inproceedings{radford2021learning,
  title={Learning transferable visual models from natural language supervision},
  author={Radford, Alec and Kim, Jong Wook and Hallacy, Chris and Ramesh, Aditya and Goh, Gabriel and Agarwal, Sandhini and Sastry, Girish and Askell, Amanda and Mishkin, Pamela and Clark, Jack and others},
  booktitle={International conference on machine learning},
  pages={8748--8763},
  year={2021},
  organization={PmLR}
}

@inproceedings{fang2022data,
  title={Data determines distributional robustness in contrastive language image pre-training (clip)},
  author={Fang, Alex and Ilharco, Gabriel and Wortsman, Mitchell and Wan, Yuhao and Shankar, Vaishaal and Dave, Achal and Schmidt, Ludwig},
  booktitle={International Conference on Machine Learning},
  pages={6216--6234},
  year={2022},
  organization={PMLR}
}

@inproceedings{koh2021wilds,
  title={Wilds: A benchmark of in-the-wild distribution shifts},
  author={Koh, Pang Wei and Sagawa, Shiori and Marklund, Henrik and Xie, Sang Michael and Zhang, Marvin and Balsubramani, Akshay and Hu, Weihua and Yasunaga, Michihiro and Phillips, Richard Lanas and Gao, Irena and others},
  booktitle={International conference on machine learning},
  pages={5637--5664},
  year={2021},
  organization={PMLR}
}

@article{zhou2022learning,
  title={Learning to prompt for vision-language models},
  author={Zhou, Kaiyang and Yang, Jingkang and Loy, Chen Change and Liu, Ziwei},
  journal={International Journal of Computer Vision},
  volume={130},
  number={9},
  pages={2337--2348},
  year={2022},
  publisher={Springer}
}

@inproceedings{zhou2022conditional,
  title={Conditional prompt learning for vision-language models},
  author={Zhou, Kaiyang and Yang, Jingkang and Loy, Chen Change and Liu, Ziwei},
  booktitle={Proceedings of the IEEE/CVF conference on computer vision and pattern recognition},
  pages={16816--16825},
  year={2022}
}

@article{zang2022unified,
  title={Unified vision and language prompt learning},
  author={Zang, Yuhang and Li, Wei and Zhou, Kaiyang and Huang, Chen and Loy, Chen Change},
  journal={arXiv preprint arXiv:2210.07225},
  year={2022}
}

@inproceedings{yao2023visual,
  title={Visual-language prompt tuning with knowledge-guided context optimization},
  author={Yao, Hantao and Zhang, Rui and Xu, Changsheng},
  booktitle={Proceedings of the IEEE/CVF conference on computer vision and pattern recognition},
  pages={6757--6767},
  year={2023}
}

@article{kather2019predicting,
  title={Predicting survival from colorectal cancer histology slides using deep learning: A retrospective multicenter study},
  author={Kather, Jakob Nikolas and Krisam, Johannes and Charoentong, Pornpimol and Luedde, Tom and Herpel, Esther and Weis, Cleo-Aron and Gaiser, Timo and Marx, Alexander and Valous, Nektarios A and Ferber, Dyke and others},
  journal={PLoS medicine},
  volume={16},
  number={1},
  pages={e1002730},
  year={2019},
  publisher={Public Library of Science San Francisco, CA USA}
}
%






\end{document}